\documentclass[10pt,conference]{IEEEtran}

\usepackage{fancyhdr}
\usepackage{cite}
\usepackage{amsmath,amssymb,amsfonts}
\usepackage{graphicx}
\usepackage{siunitx}
\usepackage{booktabs}
\usepackage{balance}
\usepackage{url}

\fancypagestyle{firstpage}{
  \fancyhead{}
  
  \fancyfoot[C]{Accepted for Proceedings of WCCI 2026, Maastricht, Netherlands. Copyright IEEE}
}



\begin{document}

\title{Graph-Based Approaches to Learning Epileptogenic Zone Localization Using Stereo-EEG Recordings}

\author{
\IEEEauthorblockN{Daniel Wendelken}
\IEEEauthorblockA{\textit{Dept. of Computer Science}\\
\textit{University of Cincinnati}\\
Cincinnati, USA\\
wendeldr@mail.uc.edu}
\and
\IEEEauthorblockN{Brian Ervin}
\IEEEauthorblockA{\textit{Dept. of Neurology}\\
\textit{Cincinnati Children's Hospital}\\
\textit{Medical Center}\\
Cincinnati, USA\\
Brian.Ervin@cchmc.org}
\and
\IEEEauthorblockN{Ravindra Arya}
\IEEEauthorblockA{\textit{Dept. of Neurology}\\
\textit{Cincinnati Children's Hospital}\\
\textit{Medical Center}\\
Cincinnati, USA\\
Ravindra.Arya@cchmc.org}
\and
\IEEEauthorblockN{Ali A. Minai}
\IEEEauthorblockA{\textit{Dept. of Electrical}\\
\textit{and Computer Engineering}\\
\textit{University of Cincinnati}\\
Cincinnati, USA\\
minaiaa@ucmail.uc.edu}
}

\maketitle

\thispagestyle{firstpage}

\begin{abstract}
The epileptogenic zone (EZ) is the brain region that generates seizures in an individual, and is the target of epilepsy surgery. Localizing the EZ from stereo-EEG (sEEG) recordings supports surgical planning, but manual interpretation is time-consuming and focuses on seizure recordings. Graphical learning models of resting-state functional connectivity among the recorded brain regions are an attractive alternative, but depend crucially on the network topology chosen for the model.

We present a controlled study of graph-based models to explore how graph topology affects EZ localization from resting-state sEEG in 40 patients. Using the same simple learnable model and leave-one-patient-out evaluation, we compare dense graphs, anatomy- and geometry-informed priors, budgeted sparsification methods, and learned sparsification, including the proposed Region-Bridge-$c$ topology. To compare graph constructions fairly, we control the number of incoming edges per node and vary graph sparsity.

At $\approx 30\%$ edge retention, Region-Bridge-$c$ achieves the highest observed mean PR-AUC ($0.371\pm0.015$; ROC-AUC $0.743\pm0.010$) while using $\approx 69\%$ fewer edges than Dense (PR-AUC $0.349\pm0.014$). Spatial-$k$ is competitive, whereas random pruning requires near-dense retention. Learned sparsification benefits from anatomical node metadata but, on average, does not surpass the best fixed prior. Across all topologies, the best choice varies by patient. These results suggest that graph construction should be evaluated explicitly rather than treated as fixed preprocessing.
\end{abstract}

\section{Introduction}
Epilepsy affects about 50 million people worldwide \cite{who_epilepsy}.
Approximately 30\% of patients have drug-resistant epilepsy \cite{kwan2010drugresistant_definition}, and for surgically eligible cases, resective or ablative intervention can substantially improve seizure outcomes \cite{wiebe2001surgery}.
Stereo-electroencephalography (sEEG) is a key technique for localizing the epileptogenic zone (EZ) \cite{gonzalezmartinez2016seeg_ez}. sEEG involves implanting multiple needle-like multi-contact depth electrodes into targeted brain regions and recording from them. Clinical interpretation of sEEG recordings is labor-intensive and focuses on seizure (ictal) recordings. Limited seizure capture and time-consuming review have driven growing interest in analyzing resting-state between-seizure (interictal) signals.

One practical approach follows from the view that epilepsy is primarily a network disorder \cite{kramer2012network}. Abnormal functional couplings underlying seizures can persist interictally and be captured by connectivity measures between implanted contacts.
Accordingly, functional connectivity graphs provide a natural representation for learning-based EZ localization.
Graph neural networks (GNNs) can learn non-linear representations over such graphs for node-level classification \cite{kipf2017gcn,gilmer2017mpnn}.

Dense functional connectivity produces fully connected graphs that mix informative pathways with weak or irrelevant interactions; dense aggregation can amplify noise and cause oversmoothing \cite{oono2020gnn}.
Sparsifying via thresholding, $k$-NN, or anatomical masks reduces computation but introduces priors that substantially alter graph statistics and downstream results \cite{vanwijk2010comparing_brain_networks}, motivating controlled comparisons under matched per-node edge budgets.

Using resting-state sEEG from 40 patients, we benchmark dense functional graphs, spatial and anatomy-informed priors, and a differentiable learned-topology module that can be projected to a hard per-node top-$k$ graph at inference, sweeping per-node incoming-edge budgets to characterize when sparsification improves over dense connectivity.

Our primary contribution is a controlled evaluation framework for matched-budget topology comparisons; within this framework, Region-Bridge-$c$, a simple anatomy+geometry prior, achieves the highest observed mean performance while retaining $\approx 30\%$ of incoming edges.
\noindent\textbf{Contributions:}
\begin{itemize}
  \item We benchmark graph topologies for resting-state sEEG EZ node classification under matched per-node incoming-edge budgets across 40 leave-one-patient-out (LOPO) folds and five random seeds.
  \item We propose and evaluate Region-Bridge-$c$, a sparse and interpretable structural prior that improves on average over dense graphs at substantially lower edge retention.
  \item We test an entmax-based learned sparsifier, projected to a hard per-node top-$k$ graph at inference, and analyze its sensitivity to realized retention and patient-level heterogeneity.
\end{itemize}
Code and data cannot be shared as both contain patient-specific information subject to confidentiality.

\section{Related Work}
\subsection{Interictal Network Features for Localization}
Pairwise connectivity features, such as phase-locking value and coherence, can be used to build functional networks that characterize interactions relevant to seizure initiation and propagation \cite{jenssen2011seizure_propagation}.
Because pairwise analyses scale rapidly with implant size and the space of interaction measures is large, automated learning-based approaches are increasingly used to summarize and interpret these networks.
Yet network statistics and their localizing relevance are highly sensitive to how connectivity is estimated and how graphs are constructed; for example, weighted versus binary representations and sparsification settings substantially change graph properties, complicating cross-patient comparisons \cite{rubinov2010complex_network_measures,vanwijk2010comparing_brain_networks}.

\subsection{Learning Over Connectivity Graphs}
Given this sensitivity, prior learning pipelines can be viewed as jointly choosing a graph construction procedure and then a predictive model on top of the resulting network.
Classical approaches often rely on handcrafted network summaries with conventional machine learning \cite{panzica2013seeg_connectivity_graphtheory,vlachos2017effective_inflow,varotto2021ez_imbalance_resampling}.
More recent work explores learned representations over connectivity graphs, including GNN-based models in related settings such as resting-state fMRI graphs and dynamic functional connectivity \cite{nandakumar2023isbi_ez_gcn_transformer}.
Recent work has applied GNNs to sEEG contact graphs for EZ/seizure onset zone (SOZ) localization using fixed graph constructions, including spatial-distance graphs over interictal recordings for surgical planning \cite{nejedly2025braincomms} and correlation-thresholded graphs over ictal recordings for SOZ localization \cite{amir2025dualtask}; in both cases the topology is not treated as an experimental variable.
Across these directions, reported performance and interpretability remain tightly coupled to the underlying graph construction and sparsification strategy, motivating controlled evaluations of topology choices and methods that can adapt or learn sparse graph structure \cite{bullmore2009complex_brain_networks,rubinov2010complex_network_measures}.

\subsection{Graph Structure Learning}
Graph neural networks typically treat the graph topology as fixed and perform message aggregation over the resulting edge set \cite{kipf2017gcn,gilmer2017mpnn}.
In epilepsy network analysis, graph construction is not merely a preprocessing step, and choices such as connectivity estimator, thresholding strategy, and use of spatial or anatomical constraints can meaningfully affect inferred network organization and subsequent prediction performance \cite{mazrooyisebdani2020tlep_graph_ml}.

Graph structure learning (GSL) instead seeks to learn edge weights or discrete connectivity patterns jointly with the predictive model \cite{franceschi2019learning_discrete}.
Since discrete edge selection is not differentiable, many approaches learn continuous edge weights and then impose sparsity through a projection step.
In this work, we use the $\alpha$-entmax transformation \cite{peters2019entmax}, a sparse alternative to softmax, to produce sparse, nonnegative edge weights and apply a hard top-$k$ projection at inference to satisfy an explicit per-node budget for comparison.

\section{Methods}

\subsection{Problem Formulation}
We treat the identification of the EZ as a binary classification problem over a graph of sEEG contacts.
For patient $p$, we define the node set $V_p = \{1, \dots, n_p\}$, where $n_p$ is the total count of contacts/channels.
The patient is represented by the graph $G_p=(V_p, E_p^{\mathrm{full}})$, containing all pairwise connections $E_p^{\mathrm{full}}=\{(u,v):u,v\in V_p,\;u\neq v\}$. Each edge can be assigned $F$ scalar features representing various statistical relations between the nodes it connects.

Let $y_v \in \{0, 1\}$ be the label for node $v$ that indicates EZ membership.
We define $\mathbf{X}_p \in \mathbb{R}^{n_p \times n_p \times F}$ as the tensor of edge features, where the entry $\mathbf{x}_{uv}$ corresponds to the directed edge from $u$ to $v$.
Different graph topologies are realized by applying a mask or weight matrix to the edge set $E_p^{\mathrm{full}}$.

\subsection{Data, Labels, and Electrode Localization}
Data acquisition, electrode localization, and EZ labeling follow our prior study \cite{wendelken2025hfo}.
For the present work, we analyzed a cohort of $40$ patients, including 21 males and 19 females, with favorable post-surgical outcomes (International League Against Epilepsy [ILAE] scores $1$--$3$) and at least one EZ-labeled contact.
Patient age was $13.3 \pm 5.6$ years (range: 2.0--21.2).
Each patient contributed approximately 5 minutes of awake resting-state sEEG (mean $5.04\pm0.02$ min) sampled at 2,048 Hz.
Patients had $132.1 \pm 31.4$ contacts (range: 75--197), of which $15.9 \pm 11.2$ (range: 3--59) were clinically labeled as EZ.
Across the cohort, this yielded 5,282 total contacts with 635 EZ contacts (12.0\%) and 4,647 non-EZ contacts (88.0\%), resulting in a class imbalance ratio of 7.3:1.
  Contacts were labeled as EZ vs.\ non-EZ based on standard clinical practice using visual analysis of ictal sEEG.
Electrode localization was performed by co-registering post-implant CT to pre-implant MRI, transforming to the Montreal Neurological Institute (MNI) standard space, and assigning each contact to anatomical parcels and derived neuroanatomical groups using FASCILE software \cite{fascile2021}.
First, a fine-grained atlas label (called \textbf{MICCAI}) \cite{neuromorphometrics2012} serves as categorical node metadata for embedding lookup.
Second, we use a coarse region label for topology priors. Starting from nine neuroanatomic groups defined in \cite{wendelken2025hfo}, we add an \texttt{unknown} category and split each group by hemisphere. This produces a 20-class label space that we refer to as \textbf{Region20}. Region20 is used for topology construction in Intra-Region and Region-Bridge-$c$, and in one learned variant it is also used as node metadata for edge scoring.

\subsection{Graph Preprocessing and Construction}
\paragraph{Graph Representation}
For each patient $p$, we constructed a directed graph from precomputed functional connectivity tensors stored in patient feature files.
Connectivity was computed after artifact rejection, notch filtering at 60\,Hz and harmonics, common average re-referencing, and segmentation into non-overlapping 0.5\,s windows ($W\approx 600$ per patient).
For each window, we used PySPI v1.1.1 to compute a comprehensive suite of pairwise connectivity measures \cite{cliff2023unifying}, yielding a windowed edge tensor of shape $(n_p,n_p,W,Q)$, where $Q=185$ is the number of potential connectivity features per electrode pair.
We then averaged across windows to obtain a graph of shape $(n_p,n_p,Q)$. 
Each feature was independently normalized within each patient using a robust z-score (median-centered, median absolute deviation [MAD]-scaled), with the median and MAD computed separately for each patient-feature pair; thus, no normalization statistics were shared across patients or LOPO folds.

\paragraph{Feature Selection}
To reduce redundancy across connectivity measures, we performed unsupervised edge-feature selection prior to training.
Using the normalized graph features, we removed features with near-zero variance and features with substantial NaN counts.
For the remaining features, we computed absolute Spearman correlations and performed complete-linkage hierarchical clustering using the distance $\delta=1-\lvert\rho\rvert$, where $\rho$ denotes the Spearman correlation.
We cut the dendrogram at $\delta\leq0.25$ to form clusters of highly correlated features, and selected one representative per cluster.
This representative was the feature with the largest sum of intra-cluster correlations, producing the final $F=63$ features used in all experiments.
This selection used only edge-feature correlations (no EZ labels) and was performed once before LOPO splitting. A leave-one-out check confirmed high stability (22/40 folds identical; Jaccard mean 0.98, min 0.88).
The retained features span phase, coherence, correlation, distance, covariance/precision, and directed/information-theoretic measures.
We converted each $(n_p,n_p,F)$ tensor to an edge list by extracting all non-diagonal entries.

\subsection{Graph Topologies}
\begin{figure}[t]
  \centering
  \includegraphics[width=0.8\columnwidth]{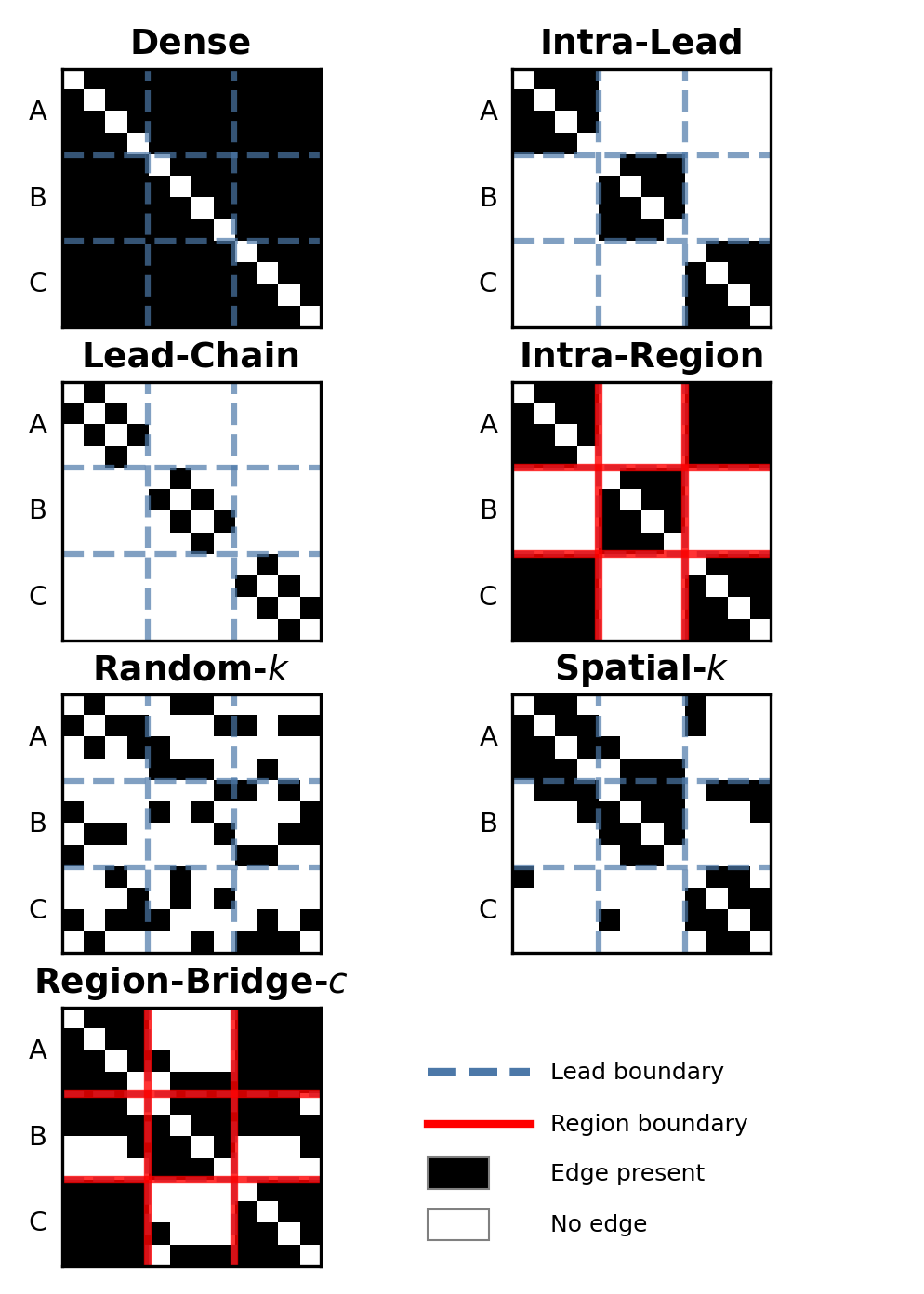}
  \caption{Adjacency-matrix schematics of evaluated topologies on a simulated 12-node sEEG graph (3 leads, 2 regions). Black entries indicate retained edges.}
  \label{fig:topology_schematic}
\end{figure}

We evaluated several topologies, illustrated schematically in Fig.~\ref{fig:topology_schematic}.

\subsubsection{\textbf{Dense}}
The dense graph configuration served as a baseline, retaining all pairwise connections except self-loops, yielding a complete directed graph.

\subsubsection{Structural Priors}
These topologies use fixed anatomical or physical constraints:
\begin{itemize}\setlength{\itemsep}{0pt}\setlength{\parskip}{0pt}
  \item \textbf{Intra-Lead:} Preserves connections between contacts on the same electrode shaft, removing inter-electrode edges.
  \item \textbf{Lead-Chain:} Restricts connectivity to adjacent contacts on the same electrode shaft.
  \item \textbf{Intra-Region:} Retains edges only between nodes sharing the same \textbf{Region20} label.
\end{itemize}

\subsubsection{Budget Controlled Topologies}
We employ budget-controlled topologies to sparsify the graph by explicitly limiting the number of incoming edges per node.
For a patient $p$ with $n_p$ contacts, each destination node $v$ retains $k=\lceil r\cdot (n_p-1)\rceil$ incoming neighbors, where $r\in(0,1]$ is the target retention ratio (fraction of possible incoming edges retained per node).
\begin{itemize}\setlength{\itemsep}{0pt}\setlength{\parskip}{0pt}
  \item \textbf{Random-$k$:} For each destination node, we sample $k$ incoming edges uniformly at random to serve as an uninformed baseline.
  \item \textbf{Spatial-$k$:} This topology connects each node to its $k$ nearest spatial neighbors, determined by Euclidean distance in 3D MNI space.
  \item \textbf{Region-Bridge-$c$:} For each node $v$, we retain all incoming edges from nodes in the same coarse anatomical region. We then add $b_v=\lceil c\,(n_p-1)\rceil$ additional incoming edges from nodes in different regions, chosen as the nearest contacts in MNI space. This hybrid topology preserves dense intra-region coupling while allowing a limited number of cross-region ``bridges.'' We set $c$ so that the realized retention ratio is close to the target retention ratio $r$; see Sec.~\ref{sec:hyperparameters}.
  \item \textbf{Learned-edges-$k$:} This approach utilizes a scoring network to predict edge importance from $\mathbf{x}_{uv}$ and applies a per-node hard top-$k$ projection during inference; see Sec.~\ref{sec:learned_topology}.
\end{itemize}

\subsection{Backbone Model}
\label{sec:backbone_model}
For each patient, the input is a directed, fully connected graph without self-loops (nodes: sEEG contacts; edges: feature vectors). The nodes and edges of the graph are processed for embedding via two distinct streams:\\
  \textbf{Node Stream:} Each node is assigned an embedding vector based on its atlas label. Each atlas label is represented by a learned vector (one vector per label, shared across contacts).\\
  \textbf{Edge Stream:} The normalized $F$-dimensional feature vector for each edge is passed through a multilayer perceptron (MLP) network to generate an edge embedding with a lower dimension.

\subsubsection{Node Label Embedding}
Each node $v$ is represented by a learnable atlas embedding $\mathbf{e}_v\in\mathbb{R}^{d_{\text{emb}}}$ (with $d_{\text{emb}}=16$), obtained by an embedding lookup indexed by the node's standardized atlas label.

\paragraph{Atlas Vocabulary Standardization:}
MICCAI labels were standardized by merging hemispheric variants to reduce label fragmentation, then frequency filtered. Labels in at least four patients receive unique IDs, and the rest map to \texttt{unknown}, producing 39 atlas IDs in this cohort that index a learnable embedding table trained end-to-end with the GNN. In the learned-topology variant, Region20 labels map without additional standardization to a separate embedding table (Sec.~\ref{sec:learned_topology_variants}).

\subsubsection{Edge Feature Embedding}
For each directed edge $(u,v)$ with edge feature vector $\mathbf{x}_{uv}\in\mathbb{R}^{F}$, we compute a node-level edge aggregation by first using an edge-level MLP $\phi_{\mathrm{edge}}:\mathbb{R}^{F}\rightarrow\mathbb{R}^{H}$:
\begin{equation}
\mathbf{m}_{uv} \;=\; \phi_{\mathrm{edge}}(\mathbf{x}_{uv}) \in \mathbb{R}^{H}
\end{equation}
with hidden dimension $H=32$ and learnable weights. Incoming projected edges for each node $v$ are aggregated via a weighted mean using nonnegative edge weights $w_{uv}$ defined by the topology. The weighted mean ensures aggregation is invariant to the number of contacts per patient. The resulting {\it edge feature vector} $h_v$ is:
\begin{equation}
\mathbf{h}_v \;=\; \frac{\sum_{u\neq v} w_{uv}\,\mathbf{m}_{uv}}{\sum_{u\neq v} w_{uv}}
\end{equation}
For all topologies, other than the learned sparsifier, $w_{uv}=1$ for retained edges and $0$ otherwise. For learned topologies, $w_{uv}$ is defined as described in Sec.~\ref{sec:learned_topology}.

The final embedding for each node $v$ is obtained by concatenating its aggregated edge feature vector with the atlas label embedding, followed by layer normalization:
\begin{equation}
\mathbf{z}_v \;=\; \mathrm{LN}\big([\mathbf{h}_v;\mathbf{e}_v]\big)
\end{equation}
A {\it classifier MLP} $\phi_{\mathrm{cls}}$ with learnable weights uses this to generate 2-class logits for the node:
\begin{equation}
\boldsymbol{\ell}_v \;=\; \phi_{\mathrm{cls}}(\mathbf{z}_v)\in\mathbb{R}^{2}
\end{equation}

All topology variants use the same two-stream backbone network; only edge masking and weights change. For fixed topology cases, there are three sets of learnable parameters: 1) The weights of the atlas label embeddings; 2) The weights of the edge embedding MLP; and 3) The weights of the classifier MLP generating logits. All parameters are trained end-to-end using the node classification loss (Eq.~\eqref{eq:focal_loss}).

\subsection{Learned Topology Module}
\label{sec:learned_topology}
For the learned topology, we compute a scalar score for each directed edge $(u,v)$ from its edge feature vector $\mathbf{x}_{uv}$ using an MLP $\phi_{\mathrm{score}}$ with trainable weights:
\begin{equation}
s_{uv} \;=\; \phi_{\mathrm{score}}(\mathbf{x}_{uv}) \in \mathbb{R}
\end{equation}
As the graph starts fully connected, for a destination node $v$, its in-degree is $d_v = n_p-1$.
We collect the incoming scores into a vector
\begin{equation}
\mathbf{q}_v \;=\; [\,s_{u_1 v},\ldots,s_{u_{d_v} v}\,] \in \mathbb{R}^{d_v}.
\end{equation}
where $u_1,\ldots,u_{d_v}$ are the source nodes $u\neq v$ (in any fixed order).
We convert scores to sparse incoming weights per destination node using the $\alpha$-entmax transform with $\alpha=1.5$ \cite{peters2019entmax}:
\begin{equation}
\boldsymbol{w}^{\mathrm{soft}}_v \;=\; \mathrm{entmax}_{1.5}(\mathbf{q}_v),
\qquad
\sum_{u\neq v} w^{\mathrm{soft}}_{uv}=1,\;\; w^{\mathrm{soft}}_{uv}\ge 0.
\end{equation}
Here $w^{\mathrm{soft}}_{uv}$ denotes the entry of $\boldsymbol{w}^{\mathrm{soft}}_v$ for source node $u$.

During inference, we enforce the exact per-node budget by keeping the top-$k_v$ scores for each destination node $v$, where $k_v=\left\lceil r\cdot d_v\right\rceil$ and $r$ is the target retention ratio.
Let $\mathcal{K}_v$ denote the top-$k_v$ source nodes for $v$ according to the scores in $\mathbf{q}_v$. We first mask the soft weights,
\begin{equation}
\tilde{w}_{uv} \;=\;
\begin{cases}
w^{\mathrm{soft}}_{uv} & \text{if } u\in \mathcal{K}_v\\
0 & \text{otherwise,}
\end{cases}
\end{equation}
and then renormalize,
\begin{equation}
w^{\mathrm{hard}}_{uv} \;=\; \frac{\tilde{w}_{uv}}{\sum_{u'\neq v} \tilde{w}_{u'v}}.
\end{equation}

We report the realized edge retention ratio as the fraction of edges with $w^{\mathrm{hard}}_{uv}>0$ relative to $n_p(n_p-1)$.

\paragraph{Auxiliary Budget Compliance Loss}
To encourage budget compliance in the learned model, we regularize the effective in-degree induced by the soft incoming weights,
\begin{equation}
d_{\mathrm{eff}}(v) \;=\; \frac{1}{\sum_{u\neq v} \big(w^{\mathrm{soft}}_{uv}\big)^{2}},
\qquad k_v=\left\lceil r\cdot d_v\right\rceil
\end{equation}
via a log-space matching loss
\begin{equation}
\mathcal{L}_{\mathrm{aux}} = \frac{1}{N_{\mathrm{train}}} \sum_{p\in\mathcal{P}_{\mathrm{train}}}\;\sum_{v\in V_p} (\log d_{\mathrm{eff}}(v) - \log k_v)^2
\end{equation}
During training we use $w_{uv}=w^{\mathrm{soft}}_{uv}$; during evaluation we use $w_{uv}=w^{\mathrm{hard}}_{uv}$.

\paragraph{Learned Topology Variants}
\label{sec:learned_topology_variants}
We evaluate an edge-only scorer (\textbf{Learned-edges-$k$}) and node-aware variants that include source/destination embeddings.
We additionally test atlas-augmented variants that concatenate atlas embeddings to the scorer input (\textbf{Learned-MICCAI-$k$}, \textbf{-detached-$k$}, \textbf{-separate-$k$}) and a coarse Region20 variant (\textbf{Learned-Region20-$k$}). These variants performed similarly; only the best (Learned-Region20-$k$) is reported individually.
Unless stated otherwise, learned topology results use the hard top-$k$ projection at inference time.

\subsection{Experimental Protocol}
\paragraph{LOPO Cross Validation}
We evaluate model performance using leave-one-patient-out cross-validation. In each fold, one patient is held out for testing, and the remaining patients are split at the patient level into train and validation sets with an 80\%:20\% split and a seeded shuffle. Each fold is repeated five times with random seeds. For topologies that depend on randomness, such as Random-$k$, we fix a separate data seed so that the structural masks are identical across training seeds.

\paragraph{Optimization and Objectives}
We train with AdamW for up to 250 epochs and apply early stopping with a patience of 30 based on the validation loss. The main objective is the focal loss \cite{lin2017focal} over node labels, which was chosen to address class imbalance.
For each node $v$, let $\boldsymbol{\ell}_v\in\mathbb{R}^{2}$ denote the logits and
$\boldsymbol{\pi}_v=\mathrm{softmax}(\boldsymbol{\ell}_v)$ the class probabilities.
Let $y_v\in\{0,1\}$ be the ground-truth label and $\pi_{v,y_v}$ the predicted
probability of the true class.

Let $\mathcal{P}_{\mathrm{train}}$ denote the set of training patients in a fold, and
$N_{\mathrm{train}}$ the total number of
training nodes in the set. The focal loss is
\begin{equation}
\label{eq:focal_loss}
\mathcal{L}_{\mathrm{focal}} \;=\; -\frac{1}{N_{\mathrm{train}}}
\sum_{p\in\mathcal{P}_{\mathrm{train}}}\;\sum_{v\in V_p}
\big(1-\pi_{v,y_v}\big)^{\gamma}\log \pi_{v,y_v},
\end{equation}
with $\gamma=1.0$ for the reported topology comparisons. For the learned topologies, we minimize
\begin{equation}
\mathcal{L} \;=\; \mathcal{L}_{\mathrm{focal}} + \lambda_{\mathrm{aux}}\mathcal{L}_{\mathrm{aux}},
\qquad \lambda_{\mathrm{aux}}=0.02.
\end{equation}

\paragraph{Regularization and Training Stabilization}
\label{sec:regularization_training_stabilization}
We apply dropout on edge features (0.2 on $\mathbf{x}_{uv}$ before $\phi_{\mathrm{edge}}$/$\phi_{\mathrm{score}}$), classifier input (0.1 on $\mathbf{z}_v$ before $\phi_{\mathrm{cls}}$), and atlas embeddings (0.5 on $\mathbf{e}_v$ before concatenation). For learned topology, we stabilize training by freezing the scorer network for the first 10 epochs and then unfreezing it.

\begin{figure*}[t]
  \centering
  \includegraphics[width=0.6\textwidth]{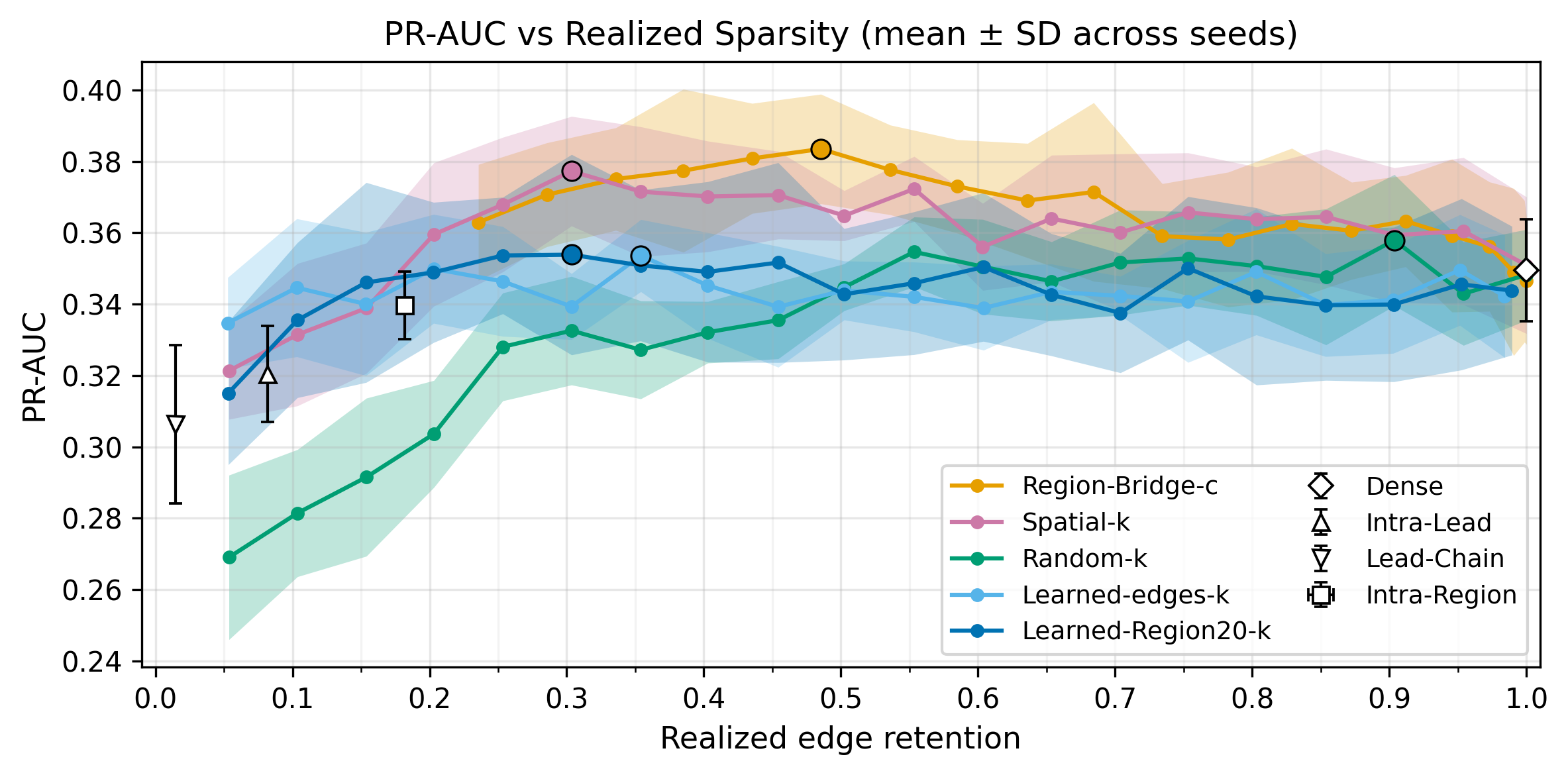}
  \caption{PR-AUC versus realized edge retention (fraction of incoming edges retained on the held-out test patients).
Lines show the mean across seeds, and error (bars and shaded bands) denote $\pm 1$ standard deviation.
Markers show the maximum observed mean PR-AUC across the sweep grid for each topology.}
  \label{fig:kratio_sweep_pr}
\end{figure*}

\paragraph{Hyperparameter Selection}
\label{sec:hyperparameters}
Backbone hyperparameters were selected in a separate grid search using a fixed baseline topology (Spatial-$k$), with validation macro precision-recall AUC (PR-AUC) as the selection metric (MCC as a tiebreaker), and then fixed for all comparisons and sweeps. Fixing hyperparameters from a single anchor topology (Spatial-$k$) is deliberate: per-topology tuning would confound topology comparisons, and Spatial-$k$ is not top-performing, so this choice does not favor the observed ranking. Unless stated otherwise, we use $H=32$, $d_{\text{emb}}=16$, $\gamma=1.0$, learning rate $0.02$ for fixed-topology models and $0.001$ for learned-topology models. Dropout probabilities are given in the Regularization and Training Stabilization paragraph above. For budgeted top-$k$ comparisons, we use target retention $r=0.3$, chosen during the initial Spatial-$k$ grid search as a moderate sparsity anchor for matched-budget comparisons; the full retention sweep is reported in Fig.~\ref{fig:kratio_sweep_pr}. All MLPs ($\phi_{\mathrm{edge}}$, $\phi_{\mathrm{cls}}$, $\phi_{\mathrm{score}}$) are 2-layer with ReLU activations.

For Region-Bridge-$c$, intra-region edges alone yield mean retention $0.182$. To match the comparison target $r=0.3$, we choose $c=0.12$, so each node adds $b_v=\left\lceil c\,(n_p-1)\right\rceil$ cross-region bridges, giving a realized retention of $0.306$.

\paragraph{Evaluation Metrics}
For each topology, metrics are computed per LOPO fold, averaged across the 40 folds within each seed, and then summarized as mean and standard deviation across the 5 seeds. To compute thresholded metrics, we choose a decision threshold on pooled validation nodes using Youden's $J$ statistic \cite{youden1950index}. This threshold is then applied to the held-out test patient to compute sensitivity, specificity, and Matthews correlation coefficient (MCC).

\paragraph{Edge Retention Reporting and Sweep}
We study sensitivity to sparsity by sweeping 20 target retention ratios
$r\in\{0.05,0.10,\ldots,1.00\}$.
For each $r$, we vary the per-node incoming-edge budget $k=\lceil r(n_p-1)\rceil$ and evaluate the resulting topology.
The sweep uses the same LOPO cross-validation protocol, training procedure, reporting, and five random seeds as the main experiments.
We report performance as a function of the mean realized edge retention on the held-out test patients.
Note that because Region-Bridge-$c$ is swept via $c$ using the same target ratio grid, its realized retentions may not precisely match those of the other topologies at the same grid point.
This is presented in Fig.~\ref{fig:kratio_sweep_pr} where the y-axis is mean PR-AUC and shaded bands denote $\pm 1$ standard deviation across seeds.

\paragraph{Ablations}
We run diagnostic ablations using the Dense topology, following the same LOPO protocol and metric computation as in the main experiments (Table~\ref{tab:ablations}).
Node-only ablations remove all edges, eliminating edge aggregation while retaining atlas node embeddings. Edge-only ablations set node embeddings to zero and retain the original edge features.
Shuffle ablations permute node metadata (atlas indices) or edge attribute vectors within each patient.

\section{Results}
We report PR-AUC as the primary metric due to strong class imbalance: 12\% EZ nodes and a random PR-AUC baseline of $\approx 0.12$. 
Table~\ref{tab:topologies} compares topologies under LOPO evaluation with a fixed backbone. 
Budget-controlled methods are compared at a matched realized edge retention of $\approx 30\%$ (target $r=0.3$) incoming edges.

\begin{table*}[t]
\centering
\small
\caption{Performance across topologies at fixed retention of 0.3 (mean \(\pm\) std). Best values are bold.}
\label{tab:topologies}

\sisetup{
  mode = text,
  detect-all,
  uncertainty-mode = compact,
  separate-uncertainty = false,
  round-mode = places,
  round-precision = 3,
  table-number-alignment = center
}

\setlength{\tabcolsep}{3pt}

\begin{tabular}{l *{6}{S[table-format=1.3(2)]}}
\toprule
Topology & {PR-AUC} & {ROC-AUC} & {MCC} & {Sens.} & {Spec.} & {Edge Retention} \\
\midrule
Dense & 0.349(14) & 0.728(17) & 0.212(11) & 0.680(26) & 0.650(31) & 1.000(00) \\
Intra-Lead & 0.320(14) & 0.691(06) & 0.171(05) & 0.602(18) & {\bfseries 0.653(17)} & 0.082(00) \\
Lead-Chain & 0.306(22) & 0.673(15) & 0.160(23) & 0.591(20) & 0.636(39) & 0.015(00) \\
Intra-Region & 0.340(09) & 0.733(13) & 0.210(16) & {\bfseries 0.694(22)} & 0.625(20) & 0.182(00) \\
\midrule
Region-Bridge-c & {\bfseries 0.371(15)} & {\bfseries 0.743(10)} & {\bfseries 0.221(11)} & 0.693(18) & 0.641(28) & 0.306(00) \\
Random-k & 0.321(13) & 0.698(02) & 0.172(08) & 0.624(40) & 0.642(21) & 0.303(00) \\
Spatial-k & 0.363(19) & 0.716(08) & 0.193(16) & 0.651(29) & 0.647(26) & 0.303(00) \\
\midrule
Learned-edges-$k$ & 0.339(09) & 0.709(14) & 0.186(16) & 0.621(45) & 0.647(20) & 0.303(00) \\
Learned-Region20-$k$ & 0.354(28) & 0.722(21) & 0.198(08) & 0.670(34) & 0.632(33) & 0.304(00) \\
\bottomrule
\end{tabular}
\end{table*}

\paragraph{Fixed Structural Priors}
Among the fixed topologies, performance varies with the restrictiveness of the prior (Table~\ref{tab:topologies}).
Intra-Region achieves the strongest PR-AUC among structural priors (0.340) and produces higher sensitivity than specificity under the Youden-selected threshold (sens. 0.694; spec. 0.625).
The more restrictive lead-based masks underperform vs. Dense. Intra-Lead achieves a PR-AUC of 0.320, and Lead-Chain achieves a PR-AUC of 0.306.
By design, the structural masks induce different graph densities (retention: Intra-Region 0.182; Intra-Lead 0.082; Lead-Chain 0.015).

\paragraph{Topology Comparisons at Matched Retention ($\approx 30\%$)}
Region-Bridge-$c$ achieves the highest observed mean PR-AUC (0.371 vs.\ 0.349 for Dense) at 0.306 retention, a 69\% edge reduction.
It also has the highest receiver operating characteristic AUC (ROC-AUC) 0.743 and MCC 0.221.
Spatial-$k$ attains 0.363 PR-AUC at 0.303 retention; Random-$k$ attains 0.321.

\paragraph{Learned Topology}
The edge-only learned sparsifier, Learned-edges-$k$, attains a PR-AUC of 0.339 and ROC-AUC of 0.709 at 0.303 retention.
The best learned variant overall, shown in Table~\ref{tab:topologies}, is Learned-Region20-$k$ (PR-AUC 0.354; ROC-AUC 0.722 at 0.304 retention). The additional atlas-augmented variants such as Learned-MICCAI-* were evaluated under matched conditions but were omitted from the table for brevity. All performed similarly, with the group-average PR-AUC at 0.345 $\pm$ 0.027 and ROC-AUC at 0.710 $\pm$ 0.015.

\paragraph{Sensitivity to Retention}
Structured priors peak at moderate retention: Region-Bridge-$c$ 0.384 (at 0.485 retention) and Spatial-$k$ 0.363 (0.303). Learned variants peak similarly: Learned-edges-$k$ 0.357 (0.354) and Learned-Region20-$k$ 0.354 (0.304). In contrast, Random-$k$ requires near-dense retention to reach its best performance (0.358 at 0.904).

\paragraph{Patient-Level Variability}
\label{par:PatVAr}
The per-patient (seed averaged) change in PR-AUC relative to Dense ranges from $-0.172$ to $+0.489$ for Region-Bridge-$c$ (mean $+0.021$).
Region-Bridge-$c$ has higher PR-AUC than Dense in 24 of 40 patients (60\%).
A Wilcoxon signed-rank test found no significant shift between Region-Bridge-$c$ and Dense (\(W=344\), \(p=0.383\)); \(r_{\mathrm{rb}}=0.161\) (95\% bootstrap CI \([-0.200,\,0.502]\)) and mean \(\Delta\)PR-AUC \(+0.021\) (95\% bootstrap CI \([-0.011,\,0.058]\)) indicate at most a weak tendency favoring Region-Bridge-$c$.
Counting the topology with the highest seed-averaged PR-AUC per patient: Learned-Region20-$k$ (7/40), Intra-Region (7/40), Learned-edges-$k$ (5/40), Dense (5/40), Lead-Chain (4/40), Region-Bridge-$c$ (4/40), Spatial-$k$ (3/40), Intra-Lead (3/40), and Random-$k$ (2/40).

\paragraph{Ablations}
Table~\ref{tab:ablations} reports ablation results.
With Dense, the edge-only ablation attains a PR-AUC of 0.351 and a ROC-AUC of 0.722, while the node-only ablation attains a PR-AUC of 0.191 and a ROC-AUC of 0.527.
Shuffling edge attributes (PR-AUC 0.162; ROC-AUC 0.506) or node metadata (PR-AUC 0.136; ROC-AUC 0.489) substantially degrades performance, indicating both contribute useful signal.
\begin{table}[t]
\centering
\footnotesize
\caption{Ablations using the Dense topology (mean $\pm$ std).}
\label{tab:ablations}

\sisetup{
  mode = text,
  detect-all,
  round-mode = places,
  round-precision = 3,
  table-number-alignment = center
}

\setlength{\tabcolsep}{3pt}
\renewcommand{\arraystretch}{1.05}

\begin{tabular}{>{\raggedright\arraybackslash}p{0.62\columnwidth} *{2}{S[table-format=1.3]}}
\toprule
Condition & {PR-AUC} & {ROC-AUC} \\
\midrule
Edge-only (no node embedding) & 0.351(18) & 0.722(15) \\
Edge-only (shuffled edge attributes) & 0.162(08) & 0.506(19) \\
Node-only (no edges) & 0.191(09) & 0.527(14) \\
Node-only (shuffled node metadata) & 0.136(02) & 0.489(06) \\
\bottomrule
\end{tabular}
\end{table}

\section{Discussion}
\paragraph{Topology choice meaningfully affects performance}
Under a fixed mean edge-aggregation backbone, topology changes the mean PR-AUC by up to 0.065 across all evaluated topologies (range 0.306--0.371; Table~\ref{tab:topologies}). Among budget-matched topologies at $\approx 30\%$ realized retention, the spread is 0.050 (0.321--0.371).
Region-Bridge-$c$ achieves the highest observed mean performance, compared with Dense (0.371 vs.\ 0.349 PR-AUC) while using $\approx 69\%$ fewer edges. Spatial-$k$ performs well, substantially outperforming Random-$k$ in PR-AUC at the same retention (0.363 vs.\ 0.321), indicating that which edges are retained matters beyond sparsity level alone.
The most restrictive structural masks (Intra-Lead and Lead-Chain) retain far fewer edges (0.082 and 0.015) and exhibit reduced discrimination (0.320 and 0.306 PR-AUC), suggesting that very local interactions and node metadata carry signal. Still, broader connectivity patterns are needed to reach the best overall performance.
Together, these results argue that graph construction is a fundamental design choice and should be reported alongside realized retention (and, when possible, sensitivity to retention) rather than treated as an unexamined preprocessing detail.

\paragraph{Why might Region-Bridge-$c$ work well?}
Region-Bridge-$c$ combines two priors: dense within-Region20 connectivity and a small set of distance-based cross-region connections.
This biases mean edge-aggregation toward anatomically coherent neighborhoods while preserving controlled inter-regional information flow; it also reduces reliance on many weak, long-range edges present in the dense graph.
The sensitivity and specificity pattern in Table~\ref{tab:topologies} is consistent with this interpretation: Intra-Region attains high sensitivity but lower specificity, while adding bridges improves specificity with similar sensitivity and yields higher overall discrimination.
This aligns with evidence that interictal abnormalities remain locally structured while seizure propagation follows fewer cross-regional pathways \cite{kramer2012network,jenssen2011seizure_propagation}.

\paragraph{Sparsification is beneficial, but not uniformly}
The retention sweep (Fig.~\ref{fig:kratio_sweep_pr}) shows that sparsification can help, but the effect depends on how edges are chosen.
Spatial-$k$ and Region-Bridge-$c$ peak at moderate retention, whereas Random-$k$ approaches its best performance near the dense graph.
Thus, different topologies have different optimal budgets, and reporting only a single sparsity level can be misleading.

\paragraph{Learned sparsification: edge-only is insufficient under hard budgets}
Under hard top-$k$ budgets, learned sparsifiers do not exceed the highest performing structural prior, Region-Bridge-$c$, at matched retention (Table~\ref{tab:topologies}).
Incorporating node metadata improves over edge-only scoring (Learned-Region20-$k$ 0.354 vs.\ Learned-edges-$k$ 0.339 PR-AUC), suggesting alignment with anatomical structure aids generalization.
Functional connectivity can be noisy and patient-specific; a globally trained scorer may overfit without contextual cues, especially in small-data settings (40 patients, LOPO).
Dense-topology ablations support this: edge attributes carry signal, but node metadata provides complementary information (Table~\ref{tab:ablations}).
In the retention sweep, learned topologies remain below the highest performing fixed priors (Fig.~\ref{fig:kratio_sweep_pr}), suggesting that hybrid approaches in which anatomy-informed priors guide learned sparsification may be more effective.

\paragraph{Patient Heterogeneity and Implications}
Patient-level results are heterogeneous across all tested topologies, indicating that topology choice can materially affect outcomes for a given implantation.
Using mean PR-AUC per patient, no single topology dominates (Sec.~\ref{par:PatVAr}).
Across patients, the median gap between the best and worst topology is 0.16 PR-AUC.
This variability likely reflects differences in implant coverage, contact counts, and the scale at which resting-state connectivity aligns with a given patient's epileptogenic network organization.
This suggests that a single global retention ratio and topology may not be optimal; patient-adaptive sparsification is promising but would require care to avoid tuning on held-out patients.

\paragraph{Limitations}
First, labels are based on clinical EZ annotations with inherent inter-rater variability; restricting to favorable ILAE outcomes reduces but does not eliminate this. The 40 patient cohort supports only within-cohort LOPO generalization; external cross-dataset and multi-center validity remain untested.
Second, window-averaged connectivity may obscure transient dynamics relevant to EZ characterization.
Third, topology priors rely on coarse region definitions and MNI-space distances; localization errors can affect both structural masks and learned metadata.
Fourth, topology rankings may depend on matched retention, a single Spatial-$k$ reference setting, and the fixed mean-aggregation backbone (Sec.~\ref{sec:backbone_model}), which we deliberately keep simple to isolate topology effects; as this model overfits without early stopping, limited capacity is unlikely the main bottleneck, but stronger architectures could still alter the rankings. We report a sweep but do not provide a principled rule for choosing $r$.
Finally, enforcing a hard top-$k$ for fair comparison may limit expressivity relative to fully weighted or adaptive schemes.
This is a controlled topology comparison, not a performance maximization study or a benchmark against clinical interpretation.

\section{Conclusions and Future Work}
We presented a controlled study of graph-topology choices for resting-state sEEG EZ localization, using a mean edge-aggregation backbone and LOPO evaluation across 40 patients.
By matching per-node incoming-edge budgets and using a shared backbone, we isolate the effect of topology from confounds caused by edge density.
Our results show that topology is a key design decision. Under a matched budget of $\approx 30\%$, Region-Bridge-$c$ achieves the highest observed mean performance (PR-AUC $0.371\pm0.015$; ROC-AUC $0.743\pm0.010$), peaking at PR-AUC $0.384$ at 48.5\% retention in a sweep. Spatial-$k$ is competitive at the same budget, while random pruning requires ${\approx}90\%$ retention to approach strong priors.
In contrast, our learned sparsifiers are competitive with structural baselines at matched retention, but they do not exceed Region-Bridge-$c$ on average.
The best sparsification varies by patient; the learned sparsifier performs best for some, reinforcing the need for patient-adaptive topology choices over a single global prior.
Finally, dense topology ablations show both edge attributes and node metadata are important, as shuffling either source degrades performance (Table~\ref{tab:ablations}).

Several directions can extend this work.
First, learned sparsifiers could incorporate stronger inductive biases that reflect successful priors, such as region-aware regularization, spatial smoothness penalties, or explicit constraints that favor sparse cross-region bridges. A practical alternative is to start from a structural prior and then fine-tune edge weights or selections with a learned module.
Second, patient-adaptive sparsity, choosing an $r$ per patient or learning a patient-specific budget, may better accommodate the heterogeneity in implants and network organization.
Third, moving beyond window-averaged static graphs to dynamic time-resolved connectivity may capture transient coupling patterns that are obscured by aggregation.
Finally, validating topology choices across multi-center cohorts, electrode layouts, and labeling practices will be necessary to assess robustness and generalization.

\section*{Acknowledgments}
We want to acknowledge support with data acquisition and clinical phenotyping from Jason Buroker, Craig Scholle, Hansel M. Greiner, Jeffrey R. Tenney, Jesse Skoch, Francesco T. Mangano, and Katherine D. Holland.
All conceptual work, methods, experimental design, data analysis, results interpretation, and the main text were produced by the authors. AI tools, specifically ChatGPT 5.2, were used for assistance in LaTeX figures, equations, table formatting, grammar review, and minor edits.

\begingroup
\footnotesize
\setlength{\itemsep}{0pt}
\balance

\endgroup

\end{document}